%% file: main.tex
\documentclass[conference]{IEEEtran}
\IEEEoverridecommandlockouts

\usepackage{cite}
\usepackage{amsmath,amssymb,amsfonts}
\usepackage{array}
\usepackage{booktabs}
\usepackage{CJKutf8}
\usepackage{multirow}
\usepackage{float}
\usepackage{graphicx}
\usepackage{tikz}
\usepackage{textcomp}
\usepackage{xcolor}
\usepackage{url}
\usepackage[hidelinks]{hyperref}

\graphicspath{{figures/}}
\input{macros}

\begin{document}
\title{Dual-Form ASR: Semantics-Aware Inverse Text Normalization for Chinese Speech Recognition}

\author{
\IEEEauthorblockN{
Fengrun Zhang\IEEEauthorrefmark{1},
Li Fu\IEEEauthorrefmark{1},
Wangjin Zhou\IEEEauthorrefmark{2},
Lu Fan\IEEEauthorrefmark{1},
Youzheng Wu\IEEEauthorrefmark{1},
Xiaodong He\IEEEauthorrefmark{1}
}
\IEEEauthorblockA{
\IEEEauthorrefmark{1}JD AI Research, Beijing, China \quad
\IEEEauthorrefmark{2}Kyoto University, Japan \\
\{zhangfengrun.1\}@jd.com
}
}

\maketitle

\begin{abstract}
\input{sections/00_abstract}

\end{abstract}


\input{sections/01_introduction}
\input{sections/02_related_work}
\input{sections/04_method}
\input{sections/05_experimental_setup}
\input{sections/06_results_and_analysis}
\input{sections/07_conclusion}


\clearpage
\bibliographystyle{IEEEtran}
\bibliography{refs/references}

\end{document}

%% file: macros.tex
\newcommand{\speechio}{\textsc{SpeechIO}}
\newcommand{\wenetspeech}{\textsc{WenetSpeech}}
\newcommand{\wetextprocessing}{\textsc{WeTextProcessing}}
\newcommand{\qwenthreefive}{Qwen3.5}
\newcommand{\vllm}{\textsc{vLLM}}
\newcommand{\fireredasr}{FireRedASR2}
\newcommand{\dualformasr}{Dual-Form ASR}
\newcommand{\dfasr}{\textsc{DF-ASR}}
\newcommand{\whisperlarge}{\textsc{Whisper-large-v3}}
\newcommand{\funasronefive}{\textsc{Fun-ASR 1.5}}
\newcommand{\doubaollmasr}{\textsc{Doubao-LLM ASR}}
\newcommand{\funasrnano}{\textsc{FunASR-Nano}}

\newcommand{\icer}{I-CER}
\newcommand{\nicer}{NI-CER}

\newcommand{\itnmwer}{ITN-MWER}
\newcommand{\fspr}{FSPR}

\newcommand{\systemcascaded}{Cascaded}
\newcommand{\systemcascadedllm}{Cascaded-LLM}

\newcommand{\systemvanillamwer}{vanilla MWER}
\newcommand{\systemkeywordmwer}{keyword-only MWER}

\newcommand{\requireitn}{\textsc{Require-ITN}}
\newcommand{\forbiditn}{\textsc{Forbid-ITN}}

\newcommand{\zh}[1]{\begin{CJK*}{UTF8}{gbsn}#1\end{CJK*}}
\newcommand{\zhsmall}[1]{{\footnotesize\zh{#1}}}

%% file: sections/00_abstract.tex
Modern automatic speech recognition (ASR) scenarios require both spoken-form transcripts for faithful transcription and readable written-form transcripts with inverse text normalization (ITN).
However, these forms are typically produced by cascaded modules, where a spoken-form ASR output is rewritten by a separate ITN component, making written-form ASR-ITN vulnerable to recognition errors and decoupling normalization from acoustic-contextual modeling, especially for semantically dependent numeric expressions.
In this paper, we propose \dualformasr{} (\dfasr{}), a framework that extends spoken-form ASR capability to semantics-aware written-form ITN through paired spoken-form and written-form supervision while retaining prompt-level selection between transcript forms.
The dual-form supervision is constructed via a large language model (LLM)-driven generate-and-judge workflow, and training is further enhanced by ITN-MWER, a sequence-level objective that assigns higher cost to errors on normalization-sensitive spans.
We also introduce a decision-aware \requireitn{}/\forbiditn{} protocol to separately measure required normalization and forbidden-span preservation.
On manually annotated Chinese subsets from \speechio{}, \dfasr{} consistently outperforms open-source ASR-ITN systems, remains competitive with strong closed-source references, and preserves reliable prompt-level control between spoken-form and written-form outputs.

%% file: sections/01_introduction.tex
\section{Introduction}

Recent advances in large language models (LLMs)~\cite{gpt4,deepseek_r1,qwen3,kimi2,glm5} are enabling automatic speech recognition (ASR) systems to generate information-rich transcripts rather than verbatim text~\cite{decoder2023jian,geng2024unveiling,fathullah2024prompting,ma2026slam}. In practical ASR deployment, readable display, captions, and meeting minutes often require written-form transcripts with inverse text normalization (ITN), whereas corpus annotation, linguistic analysis, and some downstream modules require faithful spoken-form transcription. Open Chinese ASR already provides a strong spoken-form recognition foundation, but extending this ability to written-form ASR-ITN remains difficult since ITN is not simple digit replacement: it depends on the semantic role of the expression in context.

\begin{figure}[!t]
    \centering
    \includegraphics[width=\columnwidth]{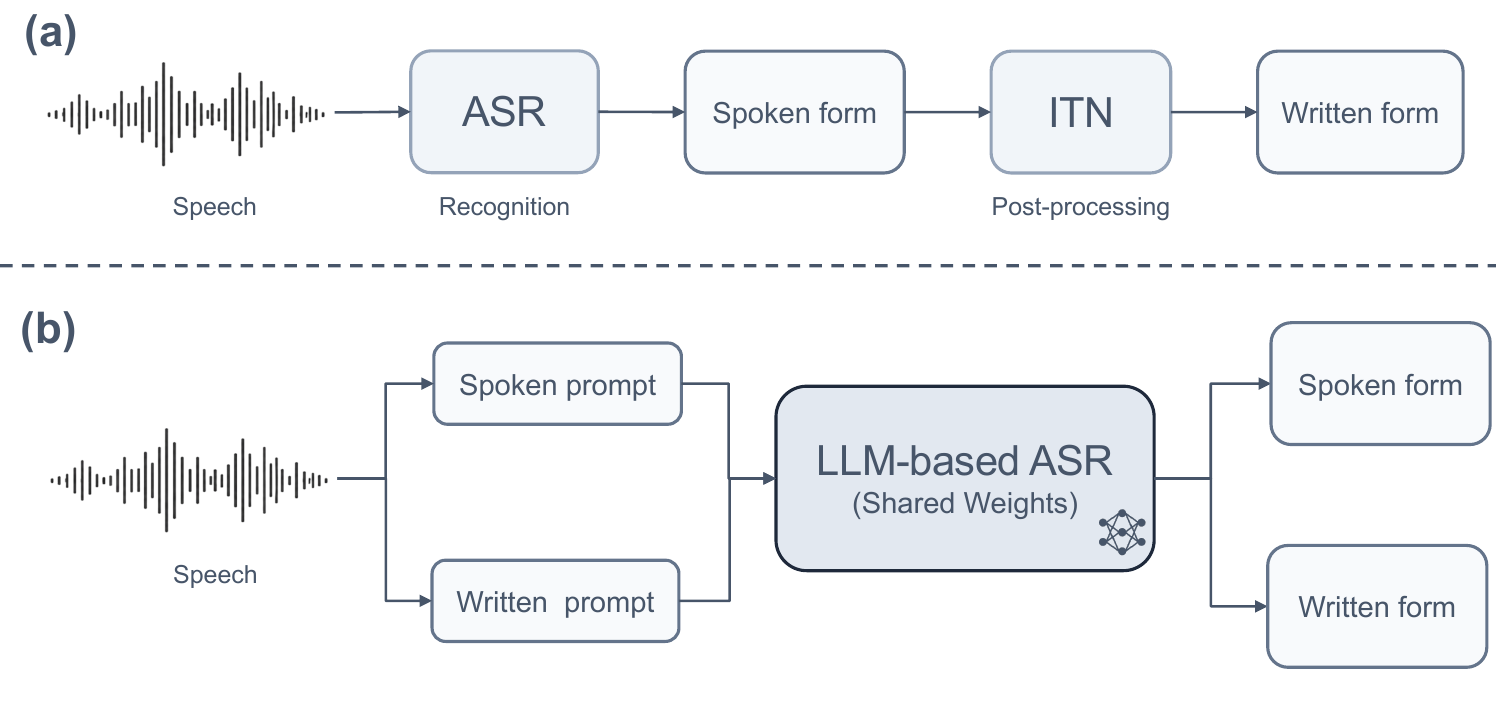}
    \caption{
    Comparison of ASR-ITN paradigms.
(a) Cascaded ASR decouples spoken-form recognition from written-form normalization by applying a separate ITN module after recognition.
(b)  \dualformasr{}  jointly trains one prompt-conditioned recognizer with paired spoken-form and written-form targets, enabling prompt-level selection between transcript forms.
}
    \label{fig:paradigm}
\end{figure}

Three gaps limit reliable open Chinese ASR-ITN. First, cascaded ITN lacks joint optimization. As shown in Fig.~\ref{fig:paradigm}~(a), practical systems usually recognize speech first and then apply a text-level ITN module~\cite{gaur2023streaming,zhang2021nemo,antonova2022thutmose,choi2024spoken,choi2025bidirectional,sunkara2021neural,tan2023four,ho2025dynamic}. The normalizer only observes recognized text, so ASR errors propagate into the ITN stage and numeric-expression normalization decisions cannot be optimized with acoustic recognition.


Second, written-form ASR-ITN is a context-sensitive semantic decision rather than a local formatting operation. The same spoken numeric expression may need to be normalized or preserved in spoken form depending on its semantic role. For example, “ten to one” may be written as “10:1” when it denotes a ratio, but should remain spoken in “it is ten to one that ...”, where the phrase means “very likely”. Such cases are difficult to maintain with rigid rule-based ITN systems, since context-dependent exceptions are hard to scale. Reliable ASR-ITN therefore requires semantics-aware normalization decisions grounded in the surrounding linguistic context.

Third, evaluation must measure over-normalization, not only required normalization. A system that aggressively converts every numeric-looking span can improve readability on ordinary quantities but damage semantically constrained expressions that should remain in spoken form. Therefore, ASR-ITN needs a protocol that jointly evaluates whether required spans are normalized and whether protected spans are preserved.

To address these gaps, we introduce \dualformasr{} (\dfasr{}), a framework that extends spoken-form ASR capability to semantics-aware written-form ITN. As illustrated in Fig.~\ref{fig:paradigm}~(b), \dfasr{} is not designed to replace spoken-form ASR or relearn acoustic recognition from scratch. Instead, the spoken-form target preserves faithful recognition behavior, while the written-form target teaches when numeric expressions should be normalized or preserved in written-form transcription. The prompt selects the transcript form, and semantic interpretation determines the numeric-expression normalization policy within written-form mode. Our contributions are summarized as follows:

\begin{itemize}
\setlength{\itemsep}{0pt}
\setlength{\parsep}{0pt}
    \item We propose \dualformasr{} (\dfasr{}), a prompt-conditioned ASR-ITN framework that extends spoken-form ASR capability to semantics-aware written-form ITN while preserving prompt-level selection between spoken-form and written-form transcripts.
    \item We construct dual-form supervision from spoken-form ASR corpora with an LLM-driven generate-and-judge workflow, and further optimize the model with ITN-MWER, an ITN-aware objective that assigns higher cost to errors on normalization-sensitive spans.
    \item We design a decision-aware \requireitn{}/\forbiditn{} evaluation protocol and show that \dfasr{} achieves 4.64\% \icer{} and 94.85\% keyword F1 on \requireitn{}, while obtaining a 95.18\% forbidden-span preservation rate (\fspr{}) on \forbiditn{}.
\end{itemize}

%% file: sections/02_related_work.tex
\section{Related Work}

\subsection{Formatted and LLM-based ASR}

LLM-based ASR combines speech encoders with powerful text decoders and has been extended to contextual biasing~\cite{yang2024mala,fu2026pac}, speaker-aware recognition~\cite{shi2026train,yin2026speakerlm}, code-switching~\cite{zhang2025boosting,liu2026cs3}, and multilingual recognition~\cite{seed_asr,qwen3_asr}. These studies show that ASR is moving beyond plain transcription toward speech-language generation with richer formatting and contextual modeling~\cite{decoder2023jian,fathullah2024prompting,ma2026slam}.

Formatted transcription can also be learned implicitly. Whisper predicts raw Internet transcripts without heavy text standardization and therefore acquires orthographic formatting abilities such as punctuation, casing, and numeric-expression normalization~\cite{radford2023whisper}. Some open LLM-based ASR systems further provide prompt interfaces for requesting formatted or normalized transcripts~\cite{an2025fun}. However, implicit formatting and prompt-elicited normalization provide only weak control over the boundary between spoken-form transcription and written-form rewriting. Without explicit supervision for both output forms, ITN behavior can become unstable and unreliable, particularly when numeric expressions require context-sensitive semantic decisions.

\subsection{Standalone ITN and Spoken-Written Conversion}

Standalone ITN is typically treated as text-level spoken-to-written conversion. Rule-based systems based on hand-crafted grammars and weighted finite-state transducers~(\mbox{WFSTs})~\cite{zhang2021nemo,wetextprocessing} are efficient on covered semiotic classes, but are difficult to maintain for open-domain, context-dependent numeric expressions. Neural converters, including sequence-to-sequence models~\cite{sunkara2021neural}, copy-or-rewrite taggers~\cite{antonova2022thutmose}, streaming transducers~\cite{gaur2023streaming,tan2023four,ho2025dynamic}, and decoder-only pretrained language models~\cite{choi2024spoken,choi2025bidirectional}, reduce rule engineering but still operate after recognition. Thus, recognition errors have already occurred, and normalization decisions cannot be optimized jointly with acoustic recognition. Since normalization is a conditional presentation decision rather than a universal action after numeric detection, these limitations motivate our dual-form setting with both faithful spoken-form and application-oriented written-form targets.

\subsection{LLM-generated Supervision and ITN Evaluation}

LLM-generated supervision and LLM-as-a-judge paradigms have been used to synthesize training signals or provide scalable quality control~\cite{wang2023selfinstruct,zheng2023judging,liu2023geval}. For ASR-ITN, however, the target is not open-ended instruction following but constrained spoken-to-written transformation: the written-form target should change appropriate numeric spans while preserving all non-ITN content.

Evaluation also needs to reflect this constrained decision. Conventional ASR metrics measure overall transcription quality, while standalone ITN metrics often emphasize whether required numeric spans are converted. They do not isolate harmful over-normalization on idioms, names, approximate quantities, and other protected spans. We therefore use LLMs to generate and judge written-form candidates, apply lightweight sanity checks for obvious invalid cases, and evaluate required normalization and forbidden-span preservation separately.

%% file: sections/04_method.tex
\section{Method}
\label{sec:method}

\dfasr{} formulates ASR-ITN as prompt-conditioned dual-form generation. Given speech $x$ and a prompt $p$, the model estimates $P_{\theta}(y\mid x,p)$. We use $p^s$ for spoken-form ASR and $p^w$ for written-form ASR-ITN. The spoken-form target preserves faithful recognition behavior, while the written-form target teaches semantics-aware ITN. The prompt selects the transcript form, and semantic interpretation determines whether spoken semiotic expressions should be normalized or preserved in written-form mode.

\subsection{LLM-driven Generate-and-Judge Dual-form Supervision}

\begin{figure}[!t]
    \centering
    \includegraphics[width=\columnwidth]{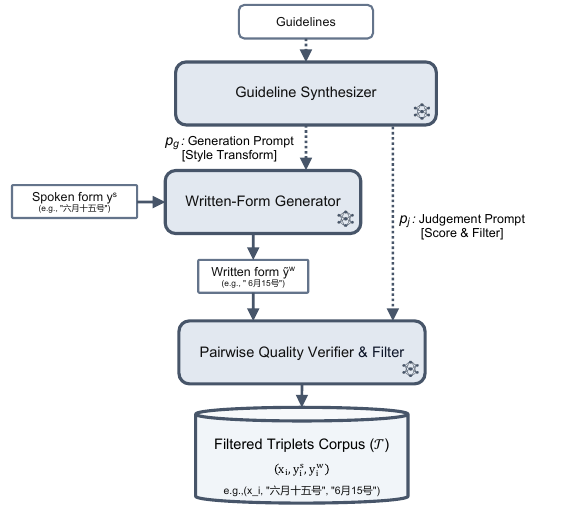}
    \caption{LLM-driven generate-and-judge workflow for dual-form supervision. An LLM generator proposes written-form candidates, an LLM verifier scores candidate quality, and lightweight deterministic checks remove obvious invalid or non-ITN rewriting artifacts.}
    \label{fig:data-agent-workflow}
\end{figure}

Fig.~\ref{fig:data-agent-workflow} illustrates the proposed LLM-driven generate-and-judge workflow for dual-form supervision. Starting from an existing ASR corpus with speech $x_i$ and a spoken-form transcript $y_i^s$, our goal is to generate a high-confidence written-form counterpart $y_i^w$. The workflow follows the general idea of LLM-generated supervision and LLM-based verification, but differs from open-ended instruction-data generation: the source transcript is fixed, and only local ITN-related spans are intended to change. LLMs are used to generate candidates and score pair quality, while lightweight deterministic sanity checks remove obvious invalid outputs before training.

First, a guideline synthesizer operationalizes the transcription standard into a generation prompt $p_g$ and a judgment prompt $p_j$, specifying how normalization-sensitive expressions should be rewritten or preserved under the written-form condition. a written-form generator $G$ is a text-only rewriter prompted to modify only ITN-related spans and propose a candidate
\begin{equation}
    \tilde{y}_i^{w}=G(y_i^{s};p_g).
\end{equation}
Second, a pairwise quality verifier  $J$ is a text-only pairwise scorer that checks whether $(y_i^s,\tilde{y}_i^w)$ preserves the original semantics, normalizes only appropriate spans, and avoids non-local rewriting:
\begin{equation}
    s_i=J(y_i^{s},\tilde{y}_i^{w};p_j).
\end{equation}
A pair is retained if the verifier score exceeds a threshold $\tau$.
Before finalizing the retained set, we apply lightweight deterministic sanity checks to remove candidates with non-ITN rewriting artifacts.
For simplicity, we denote the accepted candidate $\tilde{y}_i^w$ as $y_i^w$ and re-index the retained triples as
\begin{equation}
    \mathcal{T}=\{(x_i,y_i^s,y_i^w)\}_{i=1}^{M_T},\quad M_T=|\mathcal{T}|.
\end{equation}

This design makes the workflow auditable: the pairwise quality verifier provides scalable quality control and checks obvious invalid samples. Each retained example contributes two training instances, $(x_i,p^s,y_i^s)$ and $(x_i,p^w,y_i^w)$, so transcript form is specified by the prompt.

\subsection{Prompt-conditioned ASR-ITN Model}

\begin{figure*}[!t]
    \centering
    \makebox[\textwidth][c]{\includegraphics[width=1.10\textwidth]{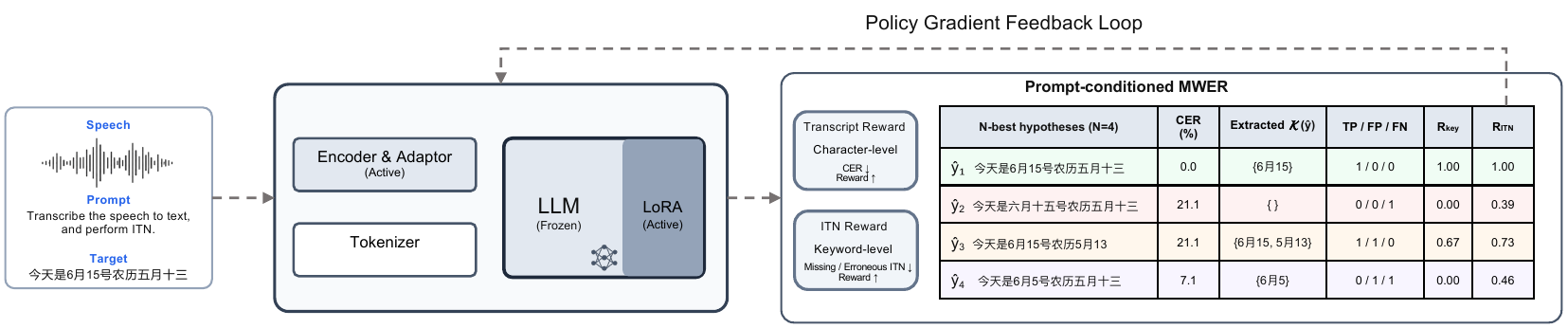}}
    \caption{\dfasr{} training with dual-form CE and ITN-MWER. ITN-MWER adds sequence-level feedback that emphasizes numeric values and units, whose errors may cause larger semantic loss than ordinary character substitutions. The figure shows 4 candidates for readability; training uses 16 candidates.}
    \label{fig:model-mwer}
\end{figure*}

As shown in Fig.~\ref{fig:model-mwer}, \dfasr{} follows an LLM-based ASR architecture with a speech encoder, an adaptor, and an LLM decoder. Acoustic representations are projected into the LLM embedding space and concatenated with textual prompt embeddings. The same model then generates either spoken-form or written-form transcripts according to the prompt.

For paired supervision $\mathcal{T}$, the dual-form cross-entropy (CE) objective is
\begin{equation}
\mathcal{L}_{\mathrm{CE}}
= -\sum_{i=1}^{M_T}\sum_{b\in\{s,w\}}\sum_{t=1}^{|y_i^b|}
    \log P_{\theta}(y_{i,t}^{b}\mid y_{i,<t}^{b},x_i,p^b).
\end{equation}

This objective prevents the model from treating spoken and written formats as noisy alternatives under the same condition. Instead, the target form is made explicit by the prompt, allowing shared acoustic recognition ability to support two deployment modes: faithful spoken-form transcription and semantics-aware written-form ASR-ITN.

\subsection{ITN-MWER for Numeric Information Loss}

While dual-form supervised fine-tuning (SFT) establishes the prompt-to-form mapping, we further refine the model with sequence-level optimization. Standard CE training optimizes token-level likelihood, and conventional minimum word error rate (MWER) training~\cite{meng2021minimum} optimizes sequence-level character errors; however, neither explicitly accounts for the unequal semantic cost of different token types. In ASR-ITN, numeric errors are often more damaging than ordinary character substitutions: dropping a decimal point, changing a percentage sign, or substituting a unit can alter the transmitted value rather than only the surface wording.

We therefore introduce ITN-MWER, a sequence-level objective that combines general transcription quality with numeric-keyword correctness. For each prompt condition $b\in\{s,w\}$, the model produces an $N$-best set $\{\hat{y}_{i,n}^{b}\}_{n=1}^{N}$ with normalized probabilities $\tilde{P}_{\theta}(\hat{y}_{i,n}^{b}\mid x_i,p^b)$. The character error rate (CER)-based reward used in vanilla MWER is

\begin{equation}
    R_{\mathrm{CER}}(\hat{y},y)=1-\mathrm{CER}(\hat{y},y).
\end{equation}

To introduce ITN-specific feedback, let $\mathcal{K}(\cdot)$ extract numeric keywords, including Arabic numbers and their attached units or symbols. Given a hypothesis $\hat{y}$ and a reference $y$, we compute matched, missing, and extra numeric keywords as TP, FN, and FP, respectively. The numeric-keyword reward is defined as
\begin{equation}
    R_{\mathrm{key}}(\hat{y},y)=\frac{2\mathrm{TP}}{2\mathrm{TP}+\mathrm{FN}+\mathrm{FP}}.
\end{equation}
If neither side contains numeric keywords, $R_{\mathrm{key}}$ is set to 1. The final ITN-MWER reward combines general transcription quality and numeric-keyword correctness:
\begin{equation}
R_{\mathrm{ITN}}(\hat{y},y)=
\frac{\alpha R_{\mathrm{CER}}(\hat{y},y)+\beta R_{\mathrm{key}}(\hat{y},y)}{\alpha+\beta}.
\end{equation}
The ITN-MWER loss uses the average reward $\bar{R}_i^b$ over the $N$-best set as a variance-reducing baseline:
\begin{equation}
\begin{aligned}
    \mathcal{L}_{\mathrm{ITN\mbox{-}MWER}}=&-\sum_{i=1}^{M_T}\sum_{b\in\{s,w\}}\sum_{n=1}^{N}
    \tilde{P}_{\theta}(\hat{y}_{i,n}^{b}\mid x_i,p^b)\\
    &\cdot\left(R_{\mathrm{ITN}}(\hat{y}_{i,n}^{b},y_i^b)-\bar{R}_{i}^{b}\right).
\end{aligned}
\end{equation}
The final objective retains the CE loss as a stabilizing term:
\begin{equation}
    \mathcal{L}
    =
    \mathcal{L}_{\mathrm{ITN\mbox{-}MWER}}
    + \lambda \mathcal{L}_{\mathrm{CE}}.
\end{equation}

The CER-based reward preserves general transcription fidelity, while the numeric-keyword reward provides ITN-specific feedback for normalization-sensitive spans.
The reward is applied under both prompt conditions so that sequence-level optimization does not collapse the dual-form behavior into a single preferred transcript style. Under the spoken-form prompt, the CER term preserves faithful transcription and prevents unnecessary rewriting. Under the written-form prompt, the numeric-keyword term provides additional feedback for normalization-sensitive spans, while the CER term constrains the model from changing unrelated context. This design aligns ITN-MWER with the dual-form objective, allowing sequence-level optimization to support written-form ASR-ITN while maintaining spoken-form transcription behavior.




%% file: sections/05_experimental_setup.tex
\section{Experimental Setup}

\subsection{Training Data and Evaluation Set}

\dfasr{} is trained from a Chinese LLM-ASR backbone using \wenetspeech{}~\cite{zhang2022wenetspeech}, covering 14.61M utterances and 9992.6 hours of speech. For dual-form supervision, we construct high-confidence targets through the proposed LLM-driven generate-and-judge workflow. The final dual-form training set contains 14.23M utterances and 9574.7 hours, where each utterance contains both a spoken-form target and a written-form target. The written-form targets include both ITN-required utterances and no-number utterances whose written targets remain identical to the spoken transcripts after basic text normalization. Among the written-form targets, 393.6K utterances, corresponding to 417.2 hours, contain Arabic numerals, accounting for 4.4\% of the written-form hours. 

For evaluation, we construct a manually annotated and reviewed Chinese ASR-ITN benchmark from \speechio{}~\cite{speechcolab_leaderboard} based on the Chinese national standard GB/T 15835--2011 for writing numerals in public texts~\cite{gbt15835_2011}. The benchmark separates two complementary deployment requirements. \requireitn{} contains 772 utterances where numeric expressions should be normalized into written form for readability. \forbiditn{} contains 309 utterances with 332 manually marked forbidden spans that should preserve spoken or lexicalized form, such as idioms, fixed expressions, proper nouns, historical terms, approximate numbers, and lexicalized numeric phrases. This split directly tests the two errors that matter in semantics-aware written-form ITN: missing required normalization and over-normalizing protected spans. We also sample 1,000 utterances without numeric spans as a no-number control set to check whether the written-form prompt affects ordinary ASR behavior or introduces false digit insertions when ITN is not required.

\subsection{Systems Compared}

We compare  \dfasr{} with cascaded systems, open-source direct ASR systems, and closed-source models. \systemcascaded{} denotes spoken-form ASR followed by \wetextprocessing{}. \systemcascadedllm{} denotes \fireredasr{} spoken-form ASR followed by a \qwenthreefive{}-35B-A3B post-processor conditioned on $p_g$. For open-source ASR-ITN references, we evaluate \funasrnano{} and \whisperlarge{}. \funasrnano{} is treated as an off-the-shelf promptable open ASR reference; its internal normalization policy and training recipe are not publicly specified, so we compare it by behavior under the same protocol rather than treating it as an architectural baseline. For closed-source models, we evaluate \doubaollmasr{}~\cite{seed_asr} and \funasronefive{}~\cite{an2025fun}. Both systems are evaluated via their public APIs on June 10, 2026. 

\input{tables/tab_require_results}

\subsection{Implementation Details}
The transcription guideline also follows the standard document GB/T 15835--2011~\cite{gbt15835_2011}. Gemini 3.0 Flash is used only to operationalize this standard into generation and judgment prompt templates. The LLM generator $G$ and verifier $J$ are both instantiated with \qwenthreefive{}-35B-A3B served by \vllm{}~\cite{kwon2023efficient}. The generator produces written-form candidates from spoken-form transcripts, while the verifier assigns a 1--10 quality score based on semantic preservation, appropriate numeric-expression normalization, and non-ITN text invariance. We set the threshold to $\tau=9$. Before finalizing retained pairs, lightweight deterministic sanity checks normalize punctuation, spaces, full-/half-width symbols, and English letter case, and remove candidates with invalid symbols or obvious non-ITN rewriting artifacts. Rejected candidates are not used as written-form supervision, since they may involve unverifiable changes beyond local numeric-expression normalization, and automatically repairing such cases would introduce additional assumptions into the training targets.

\dfasr{} is initialized with \fireredasr{}, an LLM-based ASR model~\cite{xu2026fireredasr2}, which contains a \fireredasr{} encoder, a linear adaptor with downsampling rate 2, and a Qwen2-7B-Instruct decoder. The LLM decoder is frozen except for low-rank adaptation (LoRA) adapters~\cite{hu2022lora} applied to all modules with rank 64, scaling factor 16, and dropout 0. The speech encoder and adaptor are trainable. We train with AdamW, learning rate $2\times10^{-5}$, warmup of 4,000 steps, dynamic length batching with batch size 18,000 tokens, bf16 mixed precision, gradient clipping at 5 for 1 epoch on 8 NVIDIA H200 GPUs (140GB).
To disentangle the contributions of supervision construction and sequence-level optimization, we conduct two groups of ablations. For supervision construction, we train SFT-only variants with rule-derived targets, unverified LLM-generated targets, and verified dual-form targets, respectively. These variants isolate the effect of target quality and LLM-based verification. 
We further fine-tune the verified dual-form SFT model with MWER using 16 candidates. The full ITN-MWER system sets $(\alpha,\beta)=(0.5,0.5)$, giving equal weight to the CER-based reward and the numeric-keyword reward, and sets the CE stabilizing weight to $\lambda=0.2$. We also evaluate $(\alpha,\beta)=(1,0)$ as vanilla MWER and $(\alpha,\beta)=(0,1)$ as keyword-only MWER to isolate the effect of each reward component. English letters are normalized to uppercase during evaluation to avoid case-only penalties.

\subsection{Metrics}

Following prior works~\cite{choi2024spoken,choi2025bidirectional,ho2025dynamic}, on \requireitn{} we report character error rate (CER), Inverse Character Error Rate (\icer{}), Non-Inverse Character Error Rate (\nicer{}), and numeric keyword F1. \icer{} measures errors on ITN-related regions, while \nicer{} measures non-ITN regions and is used to detect unwanted rewriting. Keyword F1 evaluates whether numeric values and units are correctly produced.

On \forbiditn{}, we report CER and forbidden-span preservation rate (\fspr{}). 
Let $N_{\mathrm{span}}^{f}$ be the total number of forbidden spans, $N_{\mathrm{itn}}^{f}$ the number incorrectly normalized into digit-bearing written forms, and $N_{\mathrm{oth}}^{f}$ the number otherwise deleted, substituted, or incompletely preserved. We define
\begin{equation}
\fspr{}=
1-\frac{N_{\mathrm{itn}}^{f}+N_{\mathrm{oth}}^{f}}
{N_{\mathrm{span}}^{f}}.
\end{equation}
A higher \fspr{} indicates better preservation of protected spoken-form spans. 

%% file: tables/tab_require_results.tex
\begin{table*}[t]
\caption{Main results on \requireitn{} and \forbiditn{}. \requireitn{} evaluates required numeric-expression normalization with \icer{}, \nicer{}, overall CER, and numeric keyword F1; \forbiditn{} evaluates protected-span preservation with CER and forbidden-span preservation rate (\fspr{}).}
\label{tab:main-results}
\centering
\footnotesize
\begin{tabular*}{0.96\textwidth}{@{\extracolsep{\fill}}lcccccc@{}}
\toprule
\multirow{2}{*}{System} &
\multicolumn{4}{c}{\requireitn{} (\%)} &
\multicolumn{2}{c}{\forbiditn{} (\%)} \\
\cmidrule(lr){2-5}\cmidrule(l){6-7}
& \icer{} $\downarrow$ & \nicer{} $\downarrow$ & CER $\downarrow$ & Keyword F1 $\uparrow$ & CER $\downarrow$ & \fspr{} $\uparrow$ \\
\midrule
\multicolumn{7}{@{}l}{\textit{Cascaded systems}} \\
\systemcascaded{} & 8.19 & 2.00 & 3.09 & 89.71 & 12.37 & 25.60 \\
\systemcascadedllm{} & 6.72 & 2.44 & 3.19 & 92.31 & 4.67 & 86.45 \\
\midrule
\multicolumn{7}{@{}l}{\textit{Open-source direct ASR systems}} \\
\funasrnano{} & 58.76 & 5.76 & 15.11 & 54.63 & 5.32 & \textbf{95.18} \\
\whisperlarge{} & 6.56 & 5.07 & 5.33 & 92.81 & 9.42 & 77.11 \\
\midrule
\multicolumn{7}{@{}l}{\textit{Closed-source models}} \\
\doubaollmasr{} & \textbf{2.47} & 2.08 & \textbf{2.15} & \textbf{95.81} & \textbf{3.56} & 93.98 \\
\funasronefive{} & 4.35 & 2.09 & 2.49 & 94.48 & 4.84 & 86.45 \\
\midrule
\multicolumn{7}{@{}l}{\textit{Proposed system}} \\
\dfasr{} & 4.64 & \textbf{1.86} & 2.35 & 94.85 & 3.67 & \textbf{95.18} \\
\bottomrule
\end{tabular*}
\vspace{0.1cm}
\end{table*}

%% file: sections/06_results_and_analysis.tex
\section{Results and Analysis}

\subsection{Required Normalization}


The \requireitn{} columns in Table~\ref{tab:main-results} evaluate required-normalization quality. Compared with the WFST cascade, \dfasr{} reduces \icer{} from 8.19\% to 4.64\%, reduces overall CER from 3.09\% to 2.35\%, and improves numeric keyword F1 from 89.71\% to 94.85\%. Compared with the LLM cascade, \dfasr{} also achieves lower \icer{} and \nicer{}, suggesting that integrated ASR-ITN is more stable than rewriting recognized text with a post-processor.

The improvement is not accompanied by increased disturbance in non-ITN regions. Among open-source direct ASR systems, \whisperlarge{} reaches competitive keyword F1 but has much worse \nicer{} than \dfasr{} (5.07\% vs. 1.86\%), indicating that its written-form output introduces more changes outside ITN-related regions. In contrast, \dfasr{} achieves strong keyword F1 while maintaining the lowest \nicer{} among all compared systems, showing a better balance between required numeric-expression normalization and surrounding transcript fidelity. Among closed-source models, \doubaollmasr{} achieves the best overall CER and keyword F1 on \requireitn{}, while \dfasr{} remains competitive with these strong external references and provides an inspectable training and evaluation recipe.

\subsection{Forbidden-span Preservation}
The \forbiditn{} columns in Table~\ref{tab:main-results} evaluate protected-span preservation. Although the LLM cascade improves \fspr{} over the WFST cascade from 25.60\% to 86.45\%, it still underperforms \dfasr{}, showing the limitation of post-processing ITN on context-dependent numeric expressions. \dfasr{} reaches 95.18\% \fspr{}, tied for the highest preservation score in the main comparison, while \doubaollmasr{} obtains the lowest CER (3.56\%) but lower \fspr{} (93.98\%). This result is noteworthy because strong industrial ASR systems may benefit from proprietary training data, product feedback, and long-term maintenance of normalization policies, whereas \dfasr{} achieves comparable forbidden-span preservation with a self-contained generate--judge--filter--train pipeline. The contrast between CER and \fspr{} also shows why preservation must be evaluated separately from ordinary transcription accuracy: a system can have low overall CER while still over-normalizing semantically protected spans.

\subsection{Required-vs-Forbidden Decision Analysis}

The two splits reveal complementary failure modes. \funasrnano{} preserves forbidden spans by avoiding many ITN decisions, reaching 95.18\% \fspr{} but only 54.63\% keyword F1 on \requireitn{}. Conversely, the WFST cascade performs many numeric conversions but over-normalizes protected spans. \dfasr{} is the most balanced open system, with 94.85\% keyword F1 on \requireitn{} and 95.18\% \fspr{} on \forbiditn{}, suggesting semantics-aware ITN rather than blind conversion or blanket preservation.

\input{tables/tab_control_set}

Table~\ref{tab:no-number-control} provides a non-regression check on a no-number control set under the written-form prompt. \dfasr{} remains close to the \fireredasr{} backbone (3.07\% vs. 2.92\% CER), and both systems introduce no false digit insertion, indicating that dual-form fine-tuning does not substantially degrade ordinary ASR behavior and that the written-form prompt does not trigger unconditional digit generation.

\subsection{Ablation Study}

\input{tables/tab_ablation_results}

Table~\ref{tab:ablation-results} separates supervision construction from sequence-level optimization. Under SFT-only training, rule-derived targets perform poorly since they inherit context-insensitive WFST decisions. Unverified LLM targets achieve the highest \fspr{}, but their \icer{} remains 6.63\%, suggesting an overly conservative behavior that preserves protected spans but still misses required normalization.Verified dual-form targets slightly lower \fspr{} but improve \icer{} and F1, showing a better balance between normalization and preservation.

For MWER objective ablation, vanilla MWER 
optimizes general character-level fidelity, while keyword-only MWER 
over-focuses on numeric tokens. By combining both rewards, ITN-MWER achieves the best \icer{} and keyword F1, showing that sequence-level optimization benefits from balancing transcription fidelity and ITN-specific feedback.

\subsection{Qualitative Analysis}

\input{tables/tab_case_study}




Table~\ref{tab:qualitative} gives dual-form examples that require context-aware normalization decisions. Cases 1 and 3 show that \dfasr{} does not simply convert every numeric-looking span. In Case 1, the first ``\zhsmall{十年}'' denotes a duration and is normalized to ``10\zhsmall{年}'', whereas the second occurs in the lexicalized expression ``\zhsmall{失去的十年}'' and is preserved. In Case 3, the Gregorian date is normalized as ``2\zhsmall{月}15\zhsmall{号}'', while the lunar-calendar expression ``\zhsmall{农历正月十一}'' remains in spoken form according to the transcription convention.

Case 2 further illustrates compositional generalization in a complex numeric statement. Although the exact decimal value ``0.9999995'' is absent from training, \dfasr{} correctly produces the decimal, the power expression, and the percentage in written form. Overall, these examples show that \dfasr{} performs prompt-controlled dual-form generation and makes context-dependent ITN decisions rather than applying local digit replacement.

%% file: tables/tab_control_set.tex
\begin{table}[t]
\caption{No-number control set results on 1,000 utterances without numeric spans.}
\label{tab:no-number-control}
\centering
\scriptsize
\setlength{\tabcolsep}{3pt}
\begin{tabular}{@{}lcc@{}}
\toprule
System & CER $\downarrow$ & False digit insertion (\%) $\downarrow$ \\
\midrule
\fireredasr{} backbone & \textbf{2.92\%} & \textbf{0.00\%} \\
\dfasr{} & 3.07\% & \textbf{0.00\%} \\
\bottomrule
\end{tabular}
\end{table}

%% file: tables/tab_ablation_results.tex
\begin{table}[t]
\caption{Ablation results for supervision sources and MWER reward settings.}
\label{tab:ablation-results}
\centering
\scriptsize
\setlength{\tabcolsep}{3pt}
\begin{tabular}{@{}lcccc@{}}
\toprule
Variant & Reward & \icer{} $\downarrow$ & Keyword F1 $\uparrow$ & \fspr{} $\uparrow$ \\
\midrule
\multicolumn{5}{@{}l}{\textit{Supervision ablation under SFT-only training}} \\
SFT w/ rule targets & -- & 8.28 & 89.17 & 26.20 \\
SFT w/ unverified targets & -- & 6.63 & 93.50 & \textbf{95.48} \\
SFT w/ verified targets & -- & 5.10 & 94.26 & 95.18 \\
\midrule
\multicolumn{5}{@{}l}{\textit{Objective ablation after verified dual-form SFT}} \\
+ \systemvanillamwer{} & $(1,0)$ & 4.87 & 94.80 & 95.18 \\
+ \systemkeywordmwer{} & $(0,1)$ & 4.96 & 94.67 & 94.88 \\
+ \itnmwer{} & $(0.5,0.5)$ & \textbf{4.64} & \textbf{94.85} & 95.18 \\
\bottomrule
\end{tabular}
\end{table}

%% file: tables/tab_case_study.tex
\begin{table}[t]
\caption{Dual-form output examples with English glosses.}
\label{tab:qualitative}
\centering
\fontsize{6.2pt}{7.0pt}\selectfont
\setlength{\tabcolsep}{1.5pt}
\renewcommand{\arraystretch}{0.80}
\begin{CJK*}{UTF8}{gbsn}
\begin{tabular}{@{}>{\scriptsize\bfseries\arraybackslash}p{0.09\columnwidth}>{\normalfont\arraybackslash}p{0.85\columnwidth}@{}}
\toprule
Cond. & {\scriptsize\textbf{Transcript}} \\
\midrule
\multicolumn{2}{@{}p{0.94\columnwidth}@{}}{{\scriptsize\textit{\textbf{Case 1.} Semantic decision: duration vs. lexicalized phrase. The first ``ten years'' is duration; the second is a lexicalized expression.}}}\\
Input & 为了使债务负担下降和经济活动恢复正常，大约需要十年或更长的时间，因此有失去的十年这种说法。\newline\textit{(To reduce the debt burden and restore economic activity, about 10 years or longer may be needed; hence the term ``the lost decade''.)}\\
$p^s$ & 为了使债务负担下降和经济活动恢复正常，大约需要十年或更长的时间，因此有失去的十年这种说法。\\
$p^w$ & 为了使债务负担下降和经济活动恢复正常，大约需要\textbf{10年}或更长的时间，因此有失去的\textbf{十年}这种说法。\\
\midrule
\multicolumn{2}{@{}p{0.94\columnwidth}@{}}{{\scriptsize\textit{\textbf{Case 2.} Unseen decimal value in a complex numeric context. The exact value 0.9999995 is absent from training but rendered correctly.}}}\\
Input & 再进行飞行，我们再去计算零点儿九九九九九九五，把它十万次幂了，这个概率是百分之九十五。\newline\textit{(For further flight calculation, we compute 0.9999995 raised to the 100,000th power; the probability is 95\%.)}\\
$p^s$ & 再进行飞行，我们再去计算零点九九九九九九五，把它十万次密了，这个概率是百分之九十五。\\
$p^w$ & 再进行飞行，我们再去计算\textbf{0.9999995}，把它\textbf{10万次}幂了，这个概率是\textbf{95\%}。\\
\midrule
\multicolumn{2}{@{}p{0.94\columnwidth}@{}}{{\scriptsize\textit{\textbf{Case 3.} Mixed decision: normalize Gregorian date, but preserve lunar date.}}}\\
Input & 今天是二月十五号，星期五，农历正月十一，欢迎收看新闻联播节目。\newline\textit{(Today is February 15, Friday, the eleventh day of the first lunar month; welcome to News Simulcast.)}\\
$p^s$ & 今天是二月十五号，星期五，农历正月十一，欢迎收看新闻联播节目。\\
$p^w$ & 今天是\textbf{2月15号}，星期五，农历正月十一，欢迎收看新闻联播节目。\\
\bottomrule
\end{tabular}
\end{CJK*}
\end{table}

%% file: sections/07_conclusion.tex
\section{Conclusion and Limitations}


We presented \dfasr{}, a Chinese ASR-ITN framework that extends spoken-form ASR capability to semantics-aware written-form ITN through paired spoken-form and written-form supervision. The LLM-driven generate-and-judge workflow constructs dual-form targets from spoken-form ASR corpora, while ITN-MWER provides sequence-level feedback for normalization-sensitive spans. The \requireitn{}/\forbiditn{} protocol separately evaluates required normalization and forbidden-span preservation. Experiments show that \dfasr{} outperforms open-source cascaded and promptable ASR-ITN baselines, remains competitive with strong closed-source references, and preserves protected spans effectively under prompt-level transcript-form control.

This work focuses on offline Chinese ASR-ITN. English or multilingual extension requires language-specific normalization conventions and evaluation sets, and the LLM-verified supervision may still contain residual numeric errors.